\documentclass[10pt,twocolumn,letterpaper]{article}

\usepackage{cvpr}              

\usepackage{graphicx}
\usepackage{amsmath}
\usepackage{amssymb}
\usepackage{booktabs}

\usepackage{multirow}
\usepackage[accsupp]{axessibility}  

\usepackage[pagebackref,breaklinks,colorlinks]{hyperref}

\usepackage[capitalize]{cleveref}
\crefname{section}{Sec.}{Secs.}
\Crefname{section}{Section}{Sections}
\Crefname{table}{Table}{Tables}
\crefname{table}{Tab.}{Tabs.}

\def\cvprPaperID{4} 
\def\confName{CVPR}
\def\confYear{2023}

\begin{document}

\title{DeepRM: Deep Recurrent Matching for 6D Pose Refinement}

\author{Alexander Avery, Andreas Savakis\\
Rochester Institute of Technology\\
Rochester, NY\\
{\tt\small \{aja9675, andreas.savakis\}@rit.edu}
}
\maketitle

\begin{abstract}

Precise 6D pose estimation of rigid objects from RGB images is a critical but challenging task in robotics, augmented reality and human-computer interaction. To address this problem, we propose DeepRM, a novel recurrent network architecture for 6D pose refinement. DeepRM leverages initial coarse pose estimates to render synthetic images of target objects. The rendered images are then matched with the observed images to predict a rigid transform for updating the previous pose estimate. This process is repeated to incrementally refine the estimate at each iteration. The DeepRM architecture incorporates LSTM units to propagate information through each refinement step, significantly improving overall performance. In contrast to current 2-stage Perspective-n-Point based solutions, DeepRM is trained end-to-end, and uses a scalable backbone that can be tuned via a single parameter for accuracy and efficiency. During training, a multi-scale optical flow head is added to predict the optical flow between the observed and synthetic images. Optical flow prediction stabilizes the training process, and enforces the learning of features that are relevant to the task of pose estimation. Our results demonstrate that DeepRM achieves state-of-the-art performance on two widely accepted challenging datasets.

\end{abstract}


\section{Introduction}

Detecting objects and estimating their 6 dimensional pose ($x$, $y$, $z$, roll, pitch, yaw) in 3D space is a fundamental task in the field of computer vision and robotics. As such, it has many applications, the most common of which is robotic manipulation. For a robot to be able to effectively interact with an object, it must know the object's pose in relation to itself. In the case of robotic grasping, the object's position is used to determine the input to the inverse kinematic solver, which can then calculate the joint states necessary to grasp the object. Augmented reality is another important field requiring very precise pose estimation \cite{Billinghurst2014a}. In this setting, pose estimation enables humans to interact with both physical and virtual objects in a seamless manner. Applications range across industries such as healthcare, manufacturing, education, and gaming.

Estimating the 6D pose from a single RGB image is an ill-posed problem due to the projection of the 3D scene onto the 2D image sensor. Because of this loss of dimensionality, many solutions rely on depth sensors to recover the depth information. Depth sensors, however, can be noisy and are typically limited by factors such as cost, power, form factor, range, resolution, frame rate, and sensitivity to external factors, e.g. sunlight \cite{Li2019a, Trabelsi2021}. Furthermore, recent advancements in computer vision and AI are enabling RGB only solutions to approach the same levels of accuracy as those with RGB-D sensors.
In the 2020 BOP Challenge on 6D Object Localization \cite{Hodan2020},
CosyPose \cite{Labb2020}, an extension of DeepIM \cite{Li2019a}, relied only on RGB data and outperformed all but two RGB-D approaches. Our intention in this work is to close the gap between RGB and RGB-D approaches by focusing on pose refinement with RGB only data, enabling our solution to be used across a wider range of applications.

\begin{figure}
\centering
\includegraphics[width=3.2in]{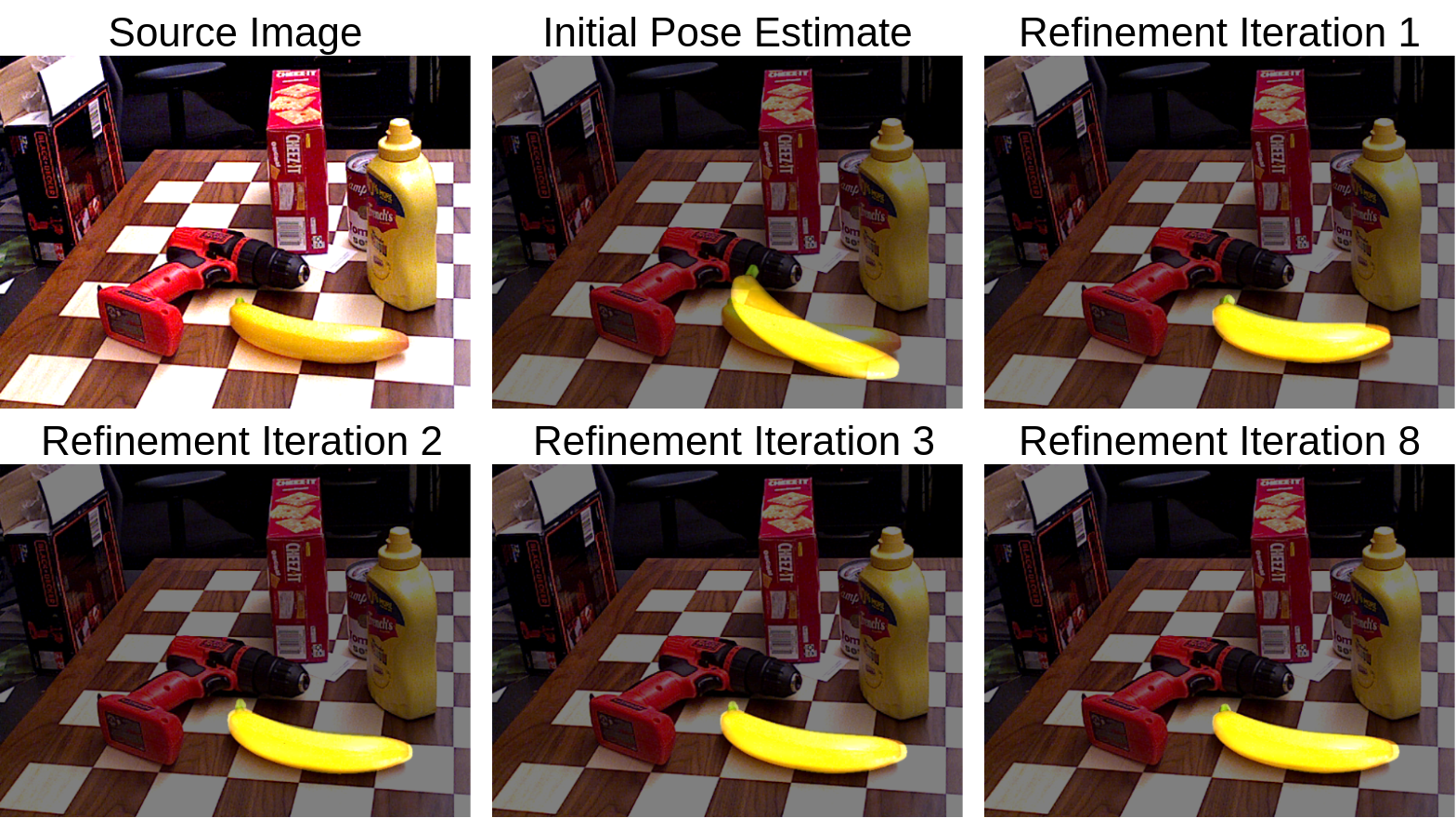}
\caption{Example of DeepRM object pose refinement. }
\label{fig:deeprm_example}
\end{figure} 

In this paper, we introduce DeepRM, a 6D pose refinement technique for rigid objects. 
Figure \ref{fig:deeprm_example} shows a representative example of the DeepRM pose refinement process.
DeepRM uses an iterative render-and-compare approach to incrementally refine an initial pose estimate. Given an initial coarse pose, a target object can be rendered with the same camera intrinsics as the original observation. The rendered image can then be matched with the observed image to predict the rigid transform that aligns the object in the two images. By leveraging the geometric information implicitly contained within the 3D model of the object, updates to the 6D pose can be inferred without external depth information.

The proposed DeepRM method improves upon DeepIM \cite{Li2019a} with several innovations, such as high resolution cropping, disentangled loss, variable renderer brightness, a scalable backbone based on EfficientNet \cite{Tan2019}, and most notably a recurrent network architecture. \textcolor{black}{
DeepRM is the first work that both leverages a recurrent neural network to directly regress 6D pose of rigid objects and provides a scalable framework for this task. Utilizing a recurrent architecture allows additional information to be propagated through each refinement step, significantly improving performance over non-recurrent methods.}

The main contributions of this paper are: 1) we present DeepRM, an end-to-end trainable recurrent neural network architecture for 6D object pose refinement, that requires only a single RGB image as input. 2) DeepRM offers a scalable solution that can be adapted based on computational constraints in real-world scenarios. 3) DeepRM achieves state-of-the-art results on the challenging YCB-Video \cite{Xiang2018a} and Occlusion LINEMOD \cite{Brachmann2014} datasets.


\begin{figure*}
\centering
\includegraphics[width=6.5in]{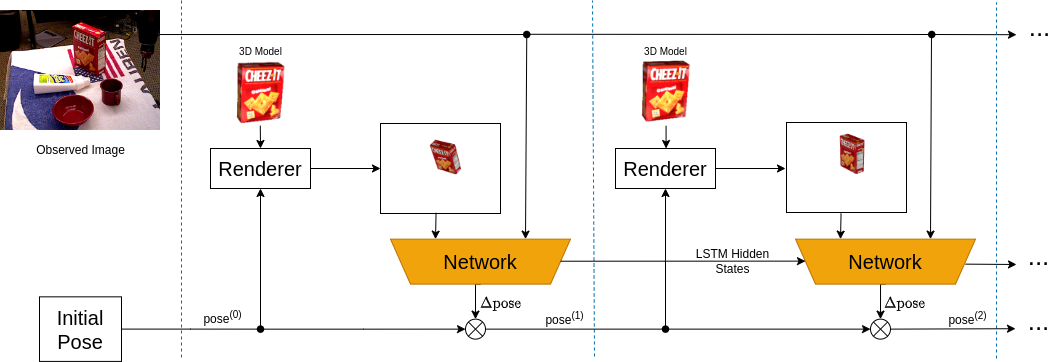}
\caption{Overview of the DeepRM method. An initial pose estimate is used to render a target object. The observed and rendered images are passed through a convolutional neural network to predict a se(3) transformation that updates the previous pose estimate. This process is repeated multiple times to incrementally refine the estimate. In addition to the updated pose estimate, hidden states from recurrent LSTM modules are propagated to each iteration.}
\label{fig:deeprm_overview}
\end{figure*} 


\section{RELATED WORK}

\subsection{6D Object Pose Estimation}

The goal of 6D object pose estimation is to determine an object's fully constrained pose within 3D space. 
As the field is vast, we limit the discussion of related works to methods based on RGB data.
Traditional methods utilized template matching techniques \cite{Hinterstoisser2012} or matched hand crafted feature points to a 3D CAD model and solved the Perspective-N-Point (PnP) problem \cite{Collet}. 
Early deep learning based methods built upon the two-stage approach of feature detection followed by PnP. BB8 \cite{Rad2017} first used this technique to regress the 8 corners of the bounding cuboid in 2D, and then solved for pose via PnP. Similar methods followed the same approach, but addressed other limiting factors such as efficiency \cite{Tekin2018} and robustness to occlusion \cite{Oberweger2018}. 

To further address the problem of occlusion, PVNet \cite{Peng2018a} introduced a pixel-wise voting network using RANSAC, resulting in an estimator that is capable of detecting keypoints, even when they are occluded.
The best results were achieved with 8 keypoints similarly to methods using bounding boxes. However, the sparsity of the keypoints in such approaches limits functionality under high levels of occlusion and truncation. To address this, a different line of research attempts to predict 3D coordinates for every pixel in the target image. By drastically increasing the number of 2D-3D correspondences, performance is maintained even under high occlusion. To handle the additional noise inherent to the dense predictions, PnP+RANSAC is needed to achieve robustness to outliers. \textcolor{black}{Dense correspondence methods include DPOD \cite{Zakharov2019}, EPOS \cite{Hodan2020a}, and ZebraPose \cite{Su2022}.}

Recent works such as PoseCNN \cite{Xiang2018a} attempt to directly regress the pose of objects from RGB images. PoseCNN uses a VGG16 \cite{Simonyan2015} backbone to extract high dimensional feature maps. These shared feature maps are then utilized by three downstream tasks: semantic segmentation to localize and distinguish objects, translation prediction, and rotation prediction. The translation and rotation predictions are directly regressed by passing flattened feature maps through fully connected layers. The benefit of direct approaches is that they can be fully trained end-to-end, without surrogate loss functions as in the two-stage approaches.

The Geometry-guided Direct Regression Network (GDR-Net) \cite{Wang2021} aims to achieve the end-to-end differentiability of direct methods, the geometry-guided accuracy of PnP methods and the robustness of dense methods.
GDR-Net  predicts dense pixel-wise correspondences, but then instead of using a non-differentiable PnP solver, it uses a a convolutional Patch-PnP network to directly regress pose. SO-Pose \cite{Di2021} further extends this approach by leveraging self occlusion information to enforce cross-layer consistencies across the correspondence field, self-occlusion information, and 6D pose, resulting in a direct method that performs comparably to many refinement based techniques.

\subsection{6D Object Pose Refinement}

Although recent methods such as GDR-Net \cite{Wang2021} and SO-Pose \cite{Di2021}, achieve high levels of accuracy compared to prior works, the ill-posedness of the problem still makes this task very challenging for RGB-only methods. As a result, refinement techniques are necessary to achieve the performance requirements of high-precision applications. Similar to traditional pose estimation techniques, early methods used either hand crafted feature descriptors, or template based matching techniques for refinement. DeepIM \cite{Li2019a} then introduced a novel neural network architecture to iteratively refine the pose of an object in a target image by matching it to a rendered image of the object's initially estimated pose. DeepIM is based on the FlowNetS \cite{Dosovitskiy2015} optical flow architecture, and directly regresses the translational and rotational updates necessary to minimize the difference in the observed and rendered images. 

Recent state-of-the-art works improve upon DeepIM by addressing a variety of factors, but virtually all of them follow the same basic render-and-compare approach. for example, CosyPose \cite{Labb2020} replaces the FlowNetS backbone with EfficientNet; \cite{Tan2019}, removes the optical flow head, and directly regresses rotation in a 6D rotation parameterization \cite{Zhou2019} as opposed to a quaternion; and 
\cite{Trabelsi2021} introduces a combined pose proposal and refinement network. Focusing on the refinement network, \cite{Trabelsi2021} extracts and warps feature maps based on the optical flow between observed and rendered images. The warped feature maps then pass through a spatial multi-attention layer to highlight important features, before directly regressing the pose update. 

RNNPose \cite{Xu2022} is a recent work on RGB pose refinement that uses an architecture inspired by RAFT \cite{Teed2020} for optical flow, but extends it significantly for the task of pose estimation. RNNPose is the first work to leverage Gated Recurrent Units (GRUs) during the iterative process of pose refinement. However, pose is optimized by a Levenberg-Marquardt (LM) algorithm on an estimated correspondence field, and therefore RNNPose is not considered a purely direct approach. 

Following RNNPose, Lipson et al. \cite{Lipson2022} also use a RAFT inspired architecture, but solve for pose using a Bidirectional Depth-Augmented PnP (BD-PnP) solver. 
This technique extends the standard PnP process by additionally minimizing the reprojection errors of the rendered image, as well as the inverse depth. Like RNNPose, this method predicts a 2D-3D correspondence field and then solves for pose, therefore we do not consider it a direct approach.

Vision transformer architectures
\cite{ViT}, \cite{Swin} 
have recently gained popularity for many computer vision tasks, including fine grained classification, semantic segmentation, object tracking, and human pose estimation. Trans6D \cite{Xu2022} and CRT-6D \cite{Castro2023} utilize vision transformers for the task of 6D object pose estimation. However, while they utilize transformers, both methods require hybridized architectures consisting of both convolutional and attention layers to achieve state-of-the-art results. CRT-6D \cite{Castro2023}, for example, uses a ResNet34 \cite{He2015DeepRL} backbone for feature extraction, followed by multiple layers of deformable self and cross-attention. Additionally, both methods require an iterative refinement process to achieve improved results.


\begin{figure*}[t]
\includegraphics[width=17.7cm]{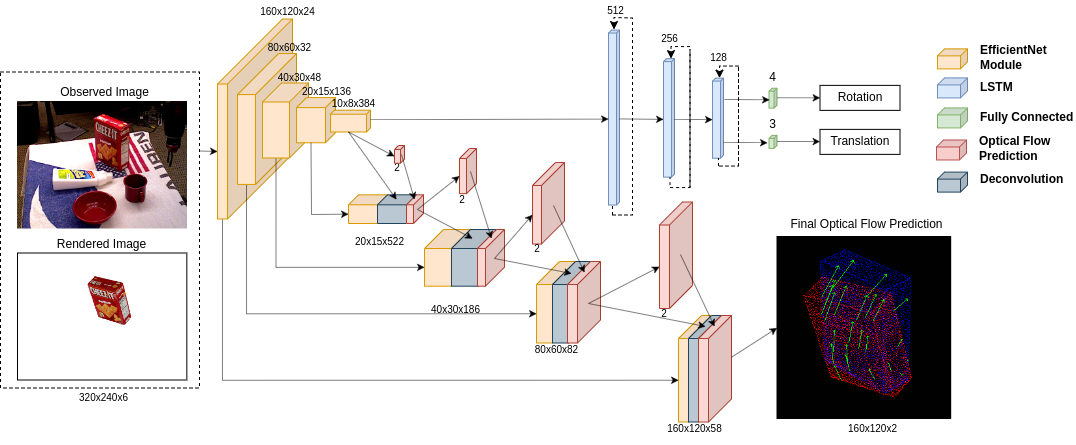}
\centering
\caption{Architecture of the proposed DeepRM method. The observed and rendered RGB images are concatenated to form a 6-channel tensor. The 6-channel tensor is then passed as input to the backbone network to extract feature maps. The final 8$\times$10$\times$384 feature map is flattened and passed through three shared, fully-connected, LSTM layers before the final translation and rotation heads. The multi-scale feature maps from the backbone network are also used in the optical flow head during training.}
\label{figure:deeprm_arch}
\end{figure*}


\section{METHOD}

An overview of the proposed DeepRM method is illustrated in Fig. \ref{fig:deeprm_overview}. Inspired by DeepIM \cite{Li2019a}, it follows an iterative render-and-compare approach to refine the pose of an object in a single RGB input image.  Given an initial pose estimate of a target object, an image of the target object is rendered. The rendered image is then matched with the real image of the object to predict an se(3) transform to the initial pose estimate that better aligns the rendered object to the observed image. The se(3) transform consists of a translation and rotation vector, where the rotation is represented as a normalized unit quaternion. se(3) denotes the Special Euclidean group, which refers to the set of proper rigid transformations within the Euclidean group. Such transforms within the Euclidean group preserve the Euclidean distance between transformed points. Because each update reduces the error between the rendered and observed images, this process can be repeated iteratively to incrementally refine the result. This method compensates for lack of external information such as depth by leveraging pre-existing geometric and visual properties of target objects, i.e. textured CAD models. By rendering objects in a way that is geometrically consistent with the observed scene, 3-D  spatial information can be recovered from the RGB only image data.

\subsection{Network Architecture}

The DeepRM neural network architecture is illustrated in Fig. \ref{figure:deeprm_arch}. The observed and rendered RGB images are concatenated channel-wise to form a 240$\times$320$\times$6 dimensional tensor. The 6-channel tensor is passed as input to the backbone convolutional neural network to extract feature maps, where the final 8$\times$10$\times$384 feature map from the backbone is flattened and passed through three shared, fully-connected, LSTM layers before the final translation and rotation heads. The multi-scale feature maps from the backbone network are also used in an auxiliary optical flow head during training to stabilize the training process, and enforce the learning of features which are relevant to the task of pose estimation. The structure of the optical flow head is the same as FlowNetS \cite{Dosovitskiy2015}, however both the spatial and channel dimensions are modified to match the corresponding layers in the backbone network. We use the B3 version of EfficientNet \cite{Tan2019} as the backbone feature extractor to achieve the best balance between performance and model size, but also demonstrate state-of-the-art results with smaller versions of EfficientNet \textcolor{black}{such as B0 and B2. Because we match the dimensions of the optical flow head to those of the backbone, our architecture can be scaled as a whole, using the same hyperparameter used by the EfficientNet backbone, called $\phi$. 
The use of $\phi$ as a model scaling hyperparameter is further discussed in \cite{Tan2019}.}

\subsection{Recurrent Fully Connected Layers}

While many other works \cite{Li2019a, Labb2020, Trabelsi2021} in pose refinement leverage an iterative process to incrementally improve upon an initial coarse estimate, most do not leverage any type of recurrent network features. However, recurrent architectures have been successfully used to improve the iterative processes of other visual processing tasks, such as optical flow prediction \cite{Teed2020}, saliency detection \cite{Deng2017}, and instance segmentation \cite{Ren2017}. Adding gated recurrent mechanisms, such as \textcolor{black}{LSTMs} or GRUs, to the iterative processes should generally maintain or improve their current levels of performance. Considering the case where all gated connections are disabled, we simply have the original network configuration, \textcolor{black}{where each iteration is independent of the previous. We can then enable the recurrent connections to enforce continuity across iterations, improving performance with each iteration.} Based on our hypothesis, we apply this theory to the task of 6D pose refinement and present a novel recurrent network architecture suited for this task.

\subsection{High Resolution Cropping}

\textcolor{black}{To improve upon the cropping strategy of DeepIM \cite{Li2019a}, we choose to follow an approach similar to  CosyPose \cite{Labb2020}. This process consists of cropping the region of interest around the object, based on the estimated pose, and then resizing this crop to 320$\times$240 before passing it to the network.} This cropping strategy has several benefits: a) it reduces background clutter b) it leverages the higher input image resolution. c) it reduces the memory and computational requirements of the network. \textcolor{black}{The only difference between our approach and \cite{Labb2020} is that we generate the rendered image at the full 640$\times$480 resolution, and use the same crop as the target image, rather than adjusting the camera parameters and rendering directly to 320$\times$240.}

\subsection{Transformation Parameterization}

Following DeepIM \cite{Li2019a}, the network does not directly predict the translational update as a vector in meters, but rather a 2D translation in pixel space, along with a relative change in depth, corresponding to the projected centerpoint of the target object. Given the initial pose of the object, and the pixel space displacements, the 3D translation can be recovered via the thin lens equation. This parameterization enables the network to perform simplified reasoning in 2D, as opposed to modeling the complex relationship between 3D object geometry and the camera intrinsics.

\subsection{Rotation Parameterization}

To regress rotation, the network predicts the four quaternion components, which are then normalized to form a unit quaternion. The advantage of normalizing the output 
is that the network 
only needs to learn
the  ratios between components.



\subsection{Disentangled Point Matching Loss}

To learn 3D pose, we use the point matching loss ($\mathcal{L}_{PML}$) function as in \cite{Li2019a}, but disentangle the translational components as in \cite{Simonelli2019}. $\mathcal{L}_{PML}$ incorporates both rotational and translational error in a single scalar metric, conveniently eliminating the need to balance the separate elements. Additionally, the disentangled formulation isolates the influence of the $xy$ translation with the relative change in depth. For a ground truth pose $\boldsymbol{p = [R|T]}$, and an estimated pose $\boldsymbol{\widetilde{p} = [\widetilde{R}|\widetilde{T}]}$, the point matching loss is defined as the average $\ell_1$ norm of a subset of n model points:
\begin{equation}
\color{black}
\mathcal{L}_{PML}(\widetilde{R},\widetilde{T}) =
\frac{1}{n}\sum\limits_{i=1}^{n}{\left|\left|(Rx_i+T)-(\widetilde{R}x_i+\widetilde{T})\right|\right|_1}.
\label{equation:pml}
\end{equation}
where $x_i$ denotes the $i{\text -}th$ model point.

Extending the above equation to disentangle the translational components, we first split the ground truth translation and the predicted translation into their respective components, i.e. $\boldsymbol{T}$ = $\boldsymbol{[x, y, z]}$ and $\boldsymbol{\widetilde{T}}$ = $\boldsymbol{[\widetilde{x}, \widetilde{y}, \widetilde{z}]}$. We then utilize a combination of the ground truth and predicted translations as input to the $\mathcal{L}_{PML}$ function to create our disentangled pose loss, $\mathcal{L}_{DPML}$:
\begin{equation}
\color{black}
\begin{split}
\mathcal{L}_{DPML} = \Big[
&\mathcal{L}_{PML}(\widetilde{R}\,,\,[\widetilde{x},\widetilde{y},\widetilde{z}]) + \\
&\mathcal{L}_{PML}(\widetilde{R}\,,\,[\widetilde{x},\widetilde{y},z]) + \\
&\mathcal{L}_{PML}(\widetilde{R}\,,\,[x,y,\widetilde{z}]) \Big] / \, 3.
\end{split}
\label{equation:disentangle}
\end{equation}

Our formulation is slightly different than \cite{Simonelli2019} and \cite{Labb2020} in that it does not disentangle the rotation component. This was found experimentally to be much more stable during training, and provides better results than the fully disentangled representation.
For the auxiliary optical flow head, we use the same multi-scale endpoint error loss ($\mathcal{L}_{MS-EPE}$) as \cite{Dosovitskiy2015}. The disentangled point matching pose loss is then combined with the mask loss to obtain the total loss ($\mathcal{L}_{total}$) as follows: 
\begin{equation}
\mathcal{L}_{total} = \mathcal{L}_{DPML} + \alpha \cdot \mathcal{L}_{MS-EPE},
\end{equation}
where the balancing factor $\alpha$ has been set to 0.1 following \cite{Li2019a}.


\section{EXPERIMENTS}

\subsection{Datasets}


The YCB-Video dataset \cite{Xiang2018a} is a a large scale dataset, with a total of 133,827 images over 92 unique scenes. Images contain labeled 6D poses of 21 target objects. The majority of images contain 4-5 objects in the scene, resulting in high levels of occlusion, as well as a variety of challenging lighting conditions. The 21 objects are a diverse selection of common household items, which include various levels of symmetry (i.e. non-symmetric, discretely symmetric, and continuously symmetric objects). For consistent comparison, we use the same exact real data, synthetic data, and data split as DeepIM \cite{Li2019a}.


The Occlusion LINEMOD dataset \cite{Brachmann2014} is an extension upon the original LINEMOD dataset \cite{Hinterstoisser2013a}. LINEMOD consists of 13 common household objects, split into 13 cluttered scenes. Roughly 1000 images are provided for each object. Many target objects are present in each image, however only a single object is labeled per image. The target object in each image is also generally very visible. To create a more challenging dataset, Occlusion LINEMOD was introduced. Occlusion LINEMOD provides ground truth labels for all objects in one of the 13 scenes. This results in high levels of partial occlusion, significantly increasing the difficulty of the dataset. Following the convention of other works such as \cite{Wang2021, Di2021}, we train on LINEMOD, and evaluate on Occlusion LINEMOD. Although, due to the limited amount of real data provided in LINEMOD, we additionally augment the training data with physically-based rendering (PBR) images that are publicly available from the 2020 BOP Challenge \cite{Hodan2020}.

\begin{table}
\centering
\begin{tabular}{ c|c|c|c }
\hline
Method& P.E.& Ref.& AUC of ADD(-S) $\uparrow$ \\
\hline
PoseCNN \cite{Xiang2018a} $\star$ & 1 & & 61.31 \\
PVNet \cite{Peng2018a} & 1 & & 73.4 \\
RePose \cite{Iwase2021} & M & \checkmark & 77.2 \\
GDR-Net \cite{Wang2021} & 1 & & 80.2 \\
DeepIM \cite{Li2019a} & 1 & \checkmark & 81.9 \\
RNNPose \cite{Xu2022} & M & \checkmark & 83.1 \\
Trabelsi \cite{Trabelsi2021} & 1 & \checkmark & 83.1 \\
SO-Pose \cite{Di2021} & 1 & & 83.9 \\
GDR-Net \cite{Wang2021} & M & & 84.4 \\
CosyPose \cite{Labb2020} & 1 & \checkmark & 84.5 \\
ZebraPose \cite{Su2022} & M & & 85.3 \\
Trans6D \cite{Xu2022} & M & \checkmark & 85.9 \\
CRT-6D \cite{Castro2023} & 1 & \checkmark & \textbf{87.5} \\
\hline
DeepRM (Ours) & 1 & \checkmark & 87.0 \\
\hline
\end{tabular}
\caption{\label{tab:ycvb_results}\textbf{Comparison to state-of-the-art on the YCB-V dataset.}\\Ref. indicates that the network includes refinement. $\star$ identifies the method used to provide initial coarse estimates to DeepRM. In the P.E. column, M indicates a separate unique model is trained per object and 1 means a single model was trained for all objects. }
\end{table}

\begin{table}
\centering
\begin{tabular}{ c|c|c|c }
\hline
Method& P.E.& Ref.& ADD(-S) 10\% $\uparrow$ \\
\hline
PoseCNN \cite{Xiang2018a} & 1 & & 24.9 \\
PVNet \cite{Peng2018a} $\star$ & 1 & & 40.8 \\
RePose \cite{Iwase2021} & M & \checkmark & 51.6 \\
PPC \cite{Brynte2020} & 1 & \checkmark & 55.3 \\
DeepIM \cite{Li2019a} & 1 & \checkmark & 55.5 \\
GDR-Net \cite{Wang2021} & 1 & & 56.1 \\
Trans6D \cite{Xu2022} & M & \checkmark & 57.9 \\
Trabelsi \cite{Trabelsi2021} & 1 & \checkmark & 58.4 \\
RNNPose \cite{Xu2022} & M & \checkmark & 60.7 \\
GDR-Net \cite{Wang2021} & M & & 62.2 \\
SO-Pose \cite{Di2021} & 1 & & 62.3 \\
CRT-6D \cite{Castro2023} & 1 & \checkmark & 66.3 \\
ZebraPose \cite{Su2022} & M & & \textbf{76.9} \\
\hline
DeepRM (Ours) & 1 & \checkmark & 65.0 \\
\hline
\end{tabular}
\caption{\label{tab:lmo_results}\textbf{Comparison to state-of-the-art on the LM-O dataset.} Ref. indicates that the network includes refinement. $\star$ identifies the method used to provide initial coarse estimates to DeepRM. In the P.E. column, M indicates a separate unique model is trained per object and 1 means a single model was trained for all objects. }
\end{table}

\subsection{Evaluation Metrics}

To evaluate the performance against other state-of-the-art methods, we follow \cite{Li2019a, Peng2018a, Labb2020, Trabelsi2021, Di2021, Wang2021} and use the ADD metric \cite{Hinterstoisser2013a}. More specifically, we use two specific variations upon it, depending on the dataset, ADD(-S) 10\% for Occlusion LINEMOD and area under the curve (AUC) ADD(-S) for YCB-Video. For the sake of brevity, we refer readers to prior works such as \cite{Hinterstoisser2013a} and \cite{Xiang2018a} for a more detailed description of these metrics.

\subsection{Implementation Details}

DeepRM is implemented in PyTorch, and uses the same OpenGL based renderer as \cite{Li2019a}. Unlike other works \cite{Li2019a, Labb2020} that use a consistent light source, we \textcolor{black}{manually tuned the renderer brightness so that it properly exposes each object that is matched in the target scene. This step can be automated by using an average metering algorithm, similar to what is used in digital photography for auto-exposure. Since the rendered object is always drawn on a black background, the render brightness for each object can be pre-determined offline, and used throughout all training and testing.} This simple modification improved the baseline results by 0.4\%. For both YCB-Video and Occlusion LINEMOD datasets, we use the ADAM optimizer \cite{Kingma2015}, with a base learning rate of 1e-4. Although due to the differences in each dataset, we use different batch sizes, training durations, and learning rate schedules for each dataset. For YCB-Video, 16 images are used per batch, with 4 objects per image, resulting in an effective batch size of 64. Similar to DeepIM \cite{Li2019a}, the model is trained for 20 epochs, with fixed learning rate decays of 0.1 at epochs 10 and 15. Although best results are obtained earlier at epoch 19 on YCB-Video. For Occlusion LINEMOD, 48 images are used per batch, with 1 object per image, resulting in an effective batch size of 48. Number of epochs are scaled up to 190, to account for the difference in the size of the dataset and batch size compared to YCB-Video. Additionally, for Occlusion LINEMOD only, a warmup period of 10\% base learning rate is used in the first 4 epochs. Both datasets are trained with 6 refinement iterations during training. Then during testing, \textcolor{black}{12} iterations of refinement are used for YCB-Video, and 6 iterations are used for Occlusion LINEMOD.


\begin{table*}
\color{black}
\centering
\setlength{\tabcolsep}{5pt}
\begin{tabular}{ c|c|c|c|c|c|c|c|c|c }
\hline
Method & Backbone & FC Type & FC Layer Dims & \# Params & Metric & 2 & 4 & 6 & 8 \\
\hline
\multirow{2}{*}{DeepIM \cite{Li2019a}} & \multirow{2}{*}{FlowNetS} & \multirow{2}{*}{MLP} & \multirow{2}{*}{256$\rightarrow$256} & \multirow{2}{*}{60M} & FPS & 12.0 & N/A & N/A & N/A \\
& & & & & ADD(-S) & N/A & 81.9 & N/A & N/A \\
\hline
\multirow{2}{*}{DeepRM (ours)} & \multirow{2}{*}{EfficientNet-B0 ($\phi$=0)} & \multirow{2}{*}{LSTM} & \multirow{2}{*}{256$\rightarrow$256$\rightarrow$128} & \multirow{2}{*}{33M} & FPS & 47.8 & 24.4 & 17.5 & 13.0 \\
& & & & & ADD(-S) & 83.2 & 84.5 & 84.7 & 84.6 \\
\hline
\multirow{2}{*}{DeepRM (ours)} & \multirow{2}{*}{EfficientNet-B2 ($\phi$=2)} & \multirow{2}{*}{LSTM} & \multirow{2}{*}{384$\rightarrow$256$\rightarrow$256} & \multirow{2}{*}{55M} & FPS & 39.2 & 21.0 & 14.0 & 10.3 \\
& & & & & ADD(-S) & 83.7 & 85.1 & 85.4 & 85.5 \\
\hline
\multirow{2}{*}{DeepRM (ours)} & \multirow{2}{*}{EfficientNet-B3 ($\phi$=3)} & \multirow{2}{*}{MLP} & \multirow{2}{*}{512$\rightarrow$256$\rightarrow$128} & \multirow{2}{*}{22M} & FPS & 37.8 & 19.5 & 13.1 & 10.1 \\
& & & & & ADD(-S) & 83.0 & 84.6 & 84.8 & 85.0 \\
\hline
\multirow{2}{*}{DeepRM (ours)} & \multirow{2}{*}{EfficientNet-B3 ($\phi$=3)} & \multirow{2}{*}{GRU} & \multirow{2}{*}{512$\rightarrow$256$\rightarrow$128} & \multirow{2}{*}{63M} & FPS & 35.7 & 18.1 & 12.7 & 9.3 \\
& & & & & ADD(-S) & 83.5 & 85.3 & 85.5 & 85.5 \\
\hline
\multirow{2}{*}{DeepRM (ours)} & \multirow{2}{*}{EfficientNet-B3 ($\phi$=3)} & \multirow{2}{*}{LSTM} & \multirow{2}{*}{512$\rightarrow$256$\rightarrow$128} & \multirow{2}{*}{79M} & FPS & 34.0 & 18.0 & 12.0 & 9.5 \\
& & & & & ADD(-S) & 84.2 & 86.2 & 86.6 & \textbf{86.8} \\
\hline
\end{tabular}
\caption{\textbf{Ablation Study on YCB-Video.} FPS represents frames per second. ADD(-S) represents AUC ADD(-S). MLP represents multi-layer perceptron, i.e. standard fully connected layer. N/A represents 'Not Available'.}
\label{tab:ycb_ablation}
\end{table*}

\begin{table*}[t] 
\color{black}
\centering
\setlength{\tabcolsep}{6pt}
\begin{tabular}{l|c|c|c|c|c|c|c|c|c|c|c|c|c|c|c}
\hline
train iters & \multirow{2}{*}{init} & \multicolumn{3}{c|}{2} & \multicolumn{4}{c|}{4} & \multicolumn{7}{c}{6} \\
\cline{1-1} 
\cline{3-16}
test iters & & 2 & 4 & 6 & 2 & 4 & 6 & 8 & 2 & 4 & 6 & 8 & 10 & 12 & 14 \\
\hline
ADD(-S) & 60.0 & 82.3 & 82.5 & 82.2 & 84.2 & 85.5 & 85.5 & 85.4 & 84.2 & 86.2 & 86.6 & 86.8 & 86.9 & \textbf{87.0} & 87.0 \\
\hline
\end{tabular}
\caption{\textbf{Ablation Study on Refinement Iterations for YCB-Video.} ADD(-S) represents AUC ADD(-S). Best results are obtained when training with 6 iterations and testing with 12.}
\label{tab:iter_ablation}
\end{table*}

\subsection{Comparison to State-of-the-Art}

\textbf{Results on YCB-Video dataset.} Table \ref{tab:ycvb_results} presents the results of DeepRM compared to the current state-of-the-art on the YCB-Video dataset for the AUC ADD(-S) metric. Initial predictions are obtained from PoseCNN \cite{Xiang2018a}, where DeepRM outperforms all existing state-of-the-art methods except CRT-6D \cite{Castro2023}. We note that CRT-6D was trained using synthetic, physically-based rendering (PBR), images for training, whereas DeepRM did not use this additional information. We speculate that re-training DeepRM on this improved dataset would close or surpass the 0.5\% performance gap.

\textbf{Results on LM-O dataset.} Table \ref{tab:lmo_results} presents the results of DeepRM compared to the current state-of-the-art on the Occlusion LINEMOD dataset for the ADD(-S) 10\% metric. Initial predictions are obtained from PVNet \cite{Peng2018a}, \textcolor{black}{where DeepRM outperforms all existing methods except for ZebraPose \cite{Su2022} and CRT-6D \cite{Castro2023}. Although ZebraPose provides significantly superior performance, it requires a different model for each object and runs at a significantly reduced frame rate compared to DeepRM.}

\subsection{\textcolor{black}{Ablation Study on YCB-Video}}
\textcolor{black}{
Table \ref{tab:ycb_ablation} displays network performance in terms of accuracy and frames per second (FPS) as a function of various backbone architectures, fully-connected layer types, fully-connected layer dimensions, and number of trainable parameters for the YCB-Video dataset.
Due to resource and time constraints, results are limited to 8 refinement iterations. 
All tests were performed on a desktop workstation with a single NVIDIA RTX 3060 GPU and an Intel i7-11700K CPU.
}

\textcolor{black}{
Highest accuracy is observed for the EfficientNet-B3 backbone using 8 refinement iterations. This configuration achieves 86.8\% at 9.5 FPS on the AUC ADD(-S) metric for YCB-Video. However, the number of refinement iterations can be decreased to 4 to achieve 18 FPS while still maintaining superior accuracy to all state-of-the-art methods.
}

\textcolor{black}{
Table \ref{tab:ycb_ablation} also demonstrates that the fully connected layers in our architecture can be scaled along with the EfficientNet backbone, using the same scaling parameter, $\phi$. Using this technique, the model can be adapted to meet real-world execution time or resource constraints. This flexibility along with the accuracy and efficiency of our method provide a solution that is well-suited to practical robotics applications.}

\textcolor{black}{
Finally, to support our claim that recurrent network features improve the performance of this task, Table \ref{tab:ycb_ablation} displays the impact of recurrent fully-connected layers compared to standard fully-connected ones.  We find that LSTMs provide a significant increase of 1.8\%, whereas GRUs provide a more moderate improvement of 0.5\% over the standard fully-connected baseline.
}

\subsection{\textcolor{black}{Ablation Study on Refinement Iterations for YCB-Video}}
\textcolor{black}{
The process of iterative refinement is heavily dependent on the number of iterations performed. As such, we investigate the impact of training and testing on a variety of refinement iterations. All tests were performed with the EfficientNet-B3 backbone on the YCB-Video dataset. AUC ADD(-S) results are reported in Table \ref{tab:iter_ablation}. Best performance is achieved when training with 6 iterations, and testing with 12 iterations, although we find 8 testing iterations to provide the best balance of accuracy and execution time.
}

\subsection{\textcolor{black}{Ablation Study on Optical Flow}}

\textcolor{black}{
In addition to the recurrent structure, the auxiliary optical flow head is one of the main features that distinguishes our work from others such as CosyPose \cite{Labb2020}. We find that the auxiliary optical flow head provides an accuracy improvement of 1.8\% on the EfficientNet-B3 backbone configuration of our network, clearly demonstrating its benefit. Furthermore, this improvement only costs a 5\% increase in parameters during training. At inference, this portion of the network is removed. Table \ref{tab:flow_ablation} displays these results using 6 training iterations, and 8 testing iterations on the YCB-Video dataset.
}

\begin{table}[hbt!] 
\color{black}
\centering
\begin{tabular}{ c|c|c }
\hline
Method & \# Params & AUC of ADD(-S) \\
\hline
No Flow & 75 M & 85.0 \\
Flow & 79 M & \textbf{86.8} \\
\hline
\end{tabular}
\caption{\textbf{Ablation Study on optical flow for YCB-Video.} Optical flow reinforcement provides a 1.8\% improvement, while only increasing the model size by 5\%.}
\label{tab:flow_ablation}
\end{table}


\section{CONCLUSIONS}

In this work, we introduce DeepRM, a novel method for precise 6D pose estimation of rigid objects from RGB only data. DeepRM improves upon existing render-and-compare approaches by leveraging several unique elements, such as an optical flow enforced learning process, an efficient and scalable backbone, and an LSTM enhanced iterative refinement mechanism.
DeepRM outperforms the majority of existing state-of-the-art methods on the challenging YCB-Video and Occlusion LINEMOD datsets.


{\small
\bibliographystyle{ieee_fullname}
\bibliography{IEEEabrv,root.bib}

\begin{thebibliography}{10}\itemsep=-1pt

\bibitem{Billinghurst2014a}
Mark Billinghurst, Adrian Clark, and Gun Lee.
\newblock {A survey of augmented reality}.
\newblock {\em Foundations and Trends in Human-Computer Interaction},
  8(2-3):73--272, 2014.

\bibitem{Brachmann2014}
Eric Brachmann, Alexander Krull, Frank Michel, Stefan Gumhold, Jamie Shotton,
  and Carsten Rother.
\newblock {Learning 6D object pose estimation using 3D object coordinates}.
\newblock In {\em Lecture Notes in Computer Science (including subseries
  Lecture Notes in Artificial Intelligence and Lecture Notes in
  Bioinformatics)}, volume 8690 LNCS, pages 536--551, 2014.

\bibitem{Brynte2020}
Lucas Brynte and Fredrik Kahl.
\newblock {Pose Proposal Critic: Robust Pose Refinement by Learning
  Reprojection Errors}.
\newblock In {\em Proceedings of the British Machine Vision Conference}, pages
  1--16, 2020.

\bibitem{Castro2023}
Pedro Castro and Tae-Kyun Kim.
\newblock Crt-6d: Fast 6d object pose estimation with cascaded refinement
  transformers.
\newblock 10 2023.

\bibitem{Collet}
Alvaro Collet and Manuel Martinez.
\newblock {MOPED: Object Recognition and Pose Estimation for Manipulation}.
\newblock {\em The International Journal of Robotics Research}, 30:1284--1306,
  2011.

\bibitem{Deng2017}
Zijun Deng, Xiaowei Hu, Lei Zhu, Xuemiao Xu, Jing Qin, Guoqiang Han, and
  Pheng-ann Heng.
\newblock {R³Net: Recurrent Residual Refinement Network for Saliency
  Detection}.
\newblock In {\em Proceedings of the Twenty-Seventh International Joint
  Conference on Artificial Intelligence, {IJCAI-18}}, pages 684--690, 2018.

\bibitem{Di2021}
Yan Di, Fabian Manhardt, Gu Wang, Xiangyang Ji, Nassir Navab, and Federico
  Tombari.
\newblock {SO-Pose: Exploiting Self-Occlusion for Direct 6D Pose Estimation}.
\newblock In {\em Proceedings of the IEEE International Conference on Computer
  Vision}, volume~1, 2021.

\bibitem{ViT}
Alexey Dosovitskiy, Lucas Beyer, Alexander Kolesnikov, Dirk Weissenborn,
  Xiaohua Zhai, Thomas Unterthiner, Mostafa~Dehghani an Matthias~Minderer,
  Georg Heigold, Sylvain Gelly, Jakob Uszkoreit, and Neil Houlsby.
\newblock An image is worth 16x16 words: Transformers for image recognition at
  scale.
\newblock {\em CoRR}, abs/2010.11929, 2020.

\bibitem{Dosovitskiy2015}
Alexey Dosovitskiy, Philipp Fischery, Eddy Ilg, Philip Hausser, Caner Hazirbas,
  Vladimir Golkov, Patrick Van~Der Smagt, Daniel Cremers, and Thomas Brox.
\newblock {FlowNet: Learning optical flow with convolutional networks}.
\newblock {\em Proceedings of the IEEE International Conference on Computer
  Vision}, 2015 Inter:2758--2766, 2015.

\bibitem{He2015DeepRL}
Kaiming He, X. Zhang, Shaoqing Ren, and Jian Sun.
\newblock Deep residual learning for image recognition.
\newblock {\em 2016 IEEE Conference on Computer Vision and Pattern Recognition
  (CVPR)}, pages 770--778, 2015.

\bibitem{Hinterstoisser2012}
Stefan Hinterstoisser, Cedric Cagniart, Slobodan Ilic, Peter Sturm, Nassir
  Navab, Pascal Fua, and Vincent Lepetit.
\newblock {Gradient response maps for real-time detection of textureless
  objects}.
\newblock {\em IEEE Transactions on Pattern Analysis and Machine Intelligence},
  34(5):876--888, 2012.

\bibitem{Hinterstoisser2013a}
Stefan Hinterstoisser, Vincent Lepetit, Slobodan Ilic, Stefan Holzer, Gary
  Bradski, Kurt Konolige, and Nassir Navab.
\newblock {Model based training, detection and pose estimation of texture-less
  3D objects in heavily cluttered scenes}.
\newblock {\em Lecture Notes in Computer Science (including subseries Lecture
  Notes in Artificial Intelligence and Lecture Notes in Bioinformatics)}, 7724
  LNCS(PART 1):548--562, 2013.

\bibitem{Hodan2020a}
Tom{\'{a}}{\v{s}} Hodaň, D{\'{a}}niel Bar{\'{a}}th, and Jiř Matas.
\newblock {EPOS: Estimating 6D pose of objects with symmetries}.
\newblock {\em Proceedings of the IEEE Computer Society Conference on Computer
  Vision and Pattern Recognition}, pages 11700--11709, 2020.

\bibitem{Hodan2020}
Tom{\'{a}}{\v{s}} Hodaň, Martin Sundermeyer, Bertram Drost, Yann Labb{\'{e}},
  Eric Brachmann, Frank Michel, Carsten Rother, and Jiř{\'{i}} Matas.
\newblock {BOP Challenge 2020 on 6D Object Localization}.
\newblock {\em Lecture Notes in Computer Science (including subseries Lecture
  Notes in Artificial Intelligence and Lecture Notes in Bioinformatics)}, 12536
  LNCS:577--594, 2020.

\bibitem{Iwase2021}
Shun Iwase, Xingyu Liu, Rawal Khirodkar, Rio Yokota, and Kris~M. Kitani.
\newblock {RePOSE: Real-Time Iterative Rendering and Refinement for 6D Object
  Pose Estimation}.
\newblock In {\em IEEE/CVF International Conference on Computer Vision}, 2021.

\bibitem{Kingma2015}
Diederik~P. Kingma and Jimmy~Lei Ba.
\newblock {Adam: A method for stochastic optimization}.
\newblock {\em 3rd International Conference on Learning Representations, ICLR
  2015 - Conference Track Proceedings}, pages 1--15, 2015.

\bibitem{Labb2020}
Yann Labb{\'{e}}, Justin Carpentier, Mathieu Aubry, and Josef Sivic.
\newblock {CosyPose : Consistent Multi-view Multi-object 6D Pose Estimation}.
\newblock In {\em European Conference on Computer Vision}, volume~2, pages
  574--591, 2020.

\bibitem{Li2019a}
Yi Li, Gu Wang, Xiangyang Ji, Yu Xiang, and Dieter Fox.
\newblock {DeepIM: Deep Iterative Matching for 6D Pose Estimation}.
\newblock {\em International Journal of Computer Vision}, 128(3):657--678, oct
  2020.

\bibitem{Lipson2022}
Lahav Lipson, Zachary Teed, Ankit Goyal, and Jia Deng.
\newblock {Coupled Iterative Refinement for 6D Multi-Object Pose Estimation}.
\newblock 2022.

\bibitem{Swin}
Ze Liu, Yutong Lin, Yue Cao, Han Hu, Yixuan Wei, Zheng Zhang, Stephen Lin, and
  Baining Guo.
\newblock Swin transformer: Hierarchical vision transformer using shifted
  windows.
\newblock {\em CoRR}, abs/2103.14030, 2021.

\bibitem{Oberweger2018}
Markus Oberweger, Mahdi Rad, and Vincent Lepetit.
\newblock {Making deep heatmaps robust to partial occlusions for 3D object pose
  estimation}.
\newblock {\em Lecture Notes in Computer Science (including subseries Lecture
  Notes in Artificial Intelligence and Lecture Notes in Bioinformatics)}, 11219
  LNCS:125--141, 2018.

\bibitem{Peng2018a}
Sida Peng, Yuan Liu, Qixing Huang, Xiaowei Zhou, and Hujun Bao.
\newblock {PVNET: Pixel-wise voting network for 6dof pose estimation}.
\newblock {\em Proceedings of the IEEE Computer Society Conference on Computer
  Vision and Pattern Recognition}, 2019-June:4556--4565, dec 2019.

\bibitem{Rad2017}
Mahdi Rad and Vincent Lepetit.
\newblock {BB8: A Scalable, Accurate, Robust to Partial Occlusion Method for
  Predicting the 3D Poses of Challenging Objects without Using Depth}.
\newblock In {\em Proceedings of the IEEE International Conference on Computer
  Vision}, volume 2017-Octob, pages 3848--3856, 2017.

\bibitem{Ren2017}
Mengye Ren and Richard~S. Zemel.
\newblock {End-to-end instance segmentation with recurrent attention}.
\newblock {\em Proceedings - 30th IEEE Conference on Computer Vision and
  Pattern Recognition, CVPR 2017}, 2017-Janua:293--301, 2017.

\bibitem{Simonelli2019}
Andrea Simonelli, Samuel~Rota Bulo, Lorenzo Porzi, Manuel Lopez-Antequera, and
  Peter Kontschieder.
\newblock {Disentangling monocular 3D object detection}.
\newblock {\em Proceedings of the IEEE International Conference on Computer
  Vision}, 2019-Octob:1991--1999, 2019.

\bibitem{Simonyan2015}
Karen Simonyan and Andrew Zisserman.
\newblock {Very deep convolutional networks for large-scale image recognition}.
\newblock {\em 3rd International Conference on Learning Representations, ICLR
  2015 - Conference Track Proceedings}, pages 1--14, 2015.

\bibitem{Su2022}
Yongzhi Su, Mahdi Saleh, Torben Fetzer, Jason Rambach, Nassir Navab, Benjamin
  Busam, Didier Stricker, and Federico Tombari.
\newblock {ZebraPose: Coarse to Fine Surface Encoding for 6DoF Object Pose
  Estimation}.
\newblock 2022.

\bibitem{Tan2019}
Mingxing Tan and Quoc~V. Le.
\newblock {EfficientNet: Rethinking model scaling for convolutional neural
  networks}.
\newblock {\em 36th International Conference on Machine Learning, ICML 2019},
  2019-June:10691--10700, 2019.

\bibitem{Teed2020}
Zachary Teed and Jia Deng.
\newblock {RAFT: Recurrent All-Pairs Field Transforms for Optical Flow}.
\newblock {\em Lecture Notes in Computer Science (including subseries Lecture
  Notes in Artificial Intelligence and Lecture Notes in Bioinformatics)}, 12347
  LNCS:402--419, 2020.

\bibitem{Tekin2018}
Bugra Tekin, Sudipta~N. Sinha, and Pascal Fua.
\newblock {Real-Time Seamless Single Shot 6D Object Pose Prediction}.
\newblock {\em Proceedings of the IEEE Computer Society Conference on Computer
  Vision and Pattern Recognition}, pages 292--301, 2018.

\bibitem{Trabelsi2021}
Ameni Trabelsi, Mohamed Chaabane, Nathaniel Blanchard, and Ross Beveridge.
\newblock {A Pose Proposal and Refinement Network for Better 6D Object Pose
  Estimation}.
\newblock In {\em IEEE Winter Conference on Applications of Computer Vision},
  pages 2381--2390, 2021.

\bibitem{Wang2021}
Gu Wang, Fabian Manhardt, Federico Tombari, and Xiangyang Ji.
\newblock {GDR-Net: Geometry-Guided Direct Regression Network for Monocular 6D
  Object Pose Estimation}.
\newblock {\em Proceedings of the IEEE Computer Society Conference on Computer
  Vision and Pattern Recognition}, pages 16611--16621, 2021.

\bibitem{Xiang2018a}
Yu Xiang, Tanner Schmidt, Venkatraman Narayanan, and Dieter Fox.
\newblock {PoseCNN: A convolutional neural network for 6D object pose
  estimation in cluttered scenes}.
\newblock {\em ArXiv}, may 2017.

\bibitem{Xu2022}
Yan Xu, Kwan-Yee Lin, Guofeng Zhang, Xiaogang Wang, and Hongsheng Li.
\newblock {RNNPose: Recurrent 6-DoF Object Pose Refinement with Robust
  Correspondence Field Estimation and Pose Optimization}.
\newblock {\em ArXiv}, 1, 2022.

\bibitem{Zakharov2019}
Sergey Zakharov, Ivan Shugurov, and Slobodan Ilic.
\newblock {DPOD: 6D pose object detector and refiner}.
\newblock In {\em Proceedings of the IEEE International Conference on Computer
  Vision}, volume 2019-Octob, pages 1941--1950, feb 2019.

\bibitem{Zhou2019}
Yi Zhou, Connelly Barnes, Jingwan Lu, Jimei Yang, and Hao Li.
\newblock {On the continuity of rotation representations in neural networks}.
\newblock In {\em Proceedings of the IEEE Computer Society Conference on
  Computer Vision and Pattern Recognition}, volume 2019-June, pages 5738--5746,
  2019.

\end{thebibliography}
}

\end{document}